\documentclass[10pt,twocolumn,letterpaper]{article}

\pdfoutput=1
\usepackage{cvpr}
\usepackage{times}
\usepackage{epsfig}
\usepackage{graphicx}
\usepackage{amsmath}
\usepackage{amssymb}
\usepackage{subfigure}
\usepackage{url}
\usepackage[font=small]{caption}
\usepackage{algorithm}
\usepackage[noend]{algpseudocode}
\usepackage{pdfpages}

\usepackage[pagebackref=true,breaklinks=true,letterpaper=true,colorlinks,bookmarks=false]{hyperref}

\cvprfinalcopy % *** Uncomment this line for the final submission

\def\cvprPaperID{1070} % *** Enter the CVPR Paper ID here

\ifcvprfinal\fi
\begin{document}

%%%%%%%%% TITLE
\title{From Images to Sentences through Scene Description Graphs \\ using Commonsense Reasoning and Knowledge}

%\and

\author{Somak Aditya\textsuperscript{$\dagger$}, Yezhou Yang{*}, Chitta Baral\textsuperscript{$\dagger$}, Cornelia Fermuller{*}  and Yiannis Aloimonos{*}\\
\textsuperscript{$\dagger$}School of Computing, Informatics and Decision Systems Engineering, Arizona State University \\
\textsuperscript{*}Department of Computer Science,  University of Maryland, College Park}
\maketitle
%\thispagestyle{empty}

%%%%%%%%% ABSTRACT
\begin{abstract}
   In this paper we propose the construction of linguistic descriptions of images. This is achieved through the extraction of scene description graphs (SDGs) from visual scenes using an automatically constructed knowledge base.  SDGs are constructed using both vision and reasoning. Specifically, commonsense reasoning\footnote{Commonsense reasoning and commonsense knowledge can be of many types \cite{Davis:2015:CRC:2817191.2701413}. Commonsense knowledge can belong to different levels of abstraction \cite{havasi2007conceptnet,Lenat:1995:CLI:219717.219745}. In this paper, we focus on capturing and reasoning based on knowledge about natural activities.} is applied on (a) detections obtained from existing perception methods on given images, (b) a ``commonsense'' knowledge base constructed using natural language processing of image annotations and (c) lexical ontological knowledge from resources such as WordNet. Amazon Mechanical Turk(AMT)-based evaluations on Flickr8k, Flickr30k and MS-COCO datasets show that in most cases, sentences auto-constructed from SDGs obtained by our method give a more “relevant” and “thorough” description of an image than a recent state-of-the-art image caption based approach. Our Image-Sentence Alignment Evaluation results are also comparable to that of the recent state-of-the art approaches.
\end{abstract}

%%%%%%%%% BODY TEXT
\section{Introduction}

``Imagine, for example, a computer that could look at an arbitrary scene, anything from a sunset at a fishing village to Grand Central Station at rush hour and produce a verbal description. This is a problem of overwhelming difficulty, relying as it does to finding solutions to both vision and language and then integrating them. I suspect that scene analysis will be one of the last cognitive tasks to be performed well by computers''.  This fifteen year old quote, attributed to A. Rosenfeld  \cite{stork1998hal}, one of the founders of the field of Computer Vision, pointed to the fundamental problem of generating semantics of visual scenes. Since then, researchers have attempted a few approaches that mostly centered on asking ``what'' and ``where'' questions about the scene in view. In this methodology, scenes are recognized by detecting the inside objects \cite{lowe1999object,dalal2005histograms,Krizhevsky2012},  objects are recognized by detecting their parts or attributes \cite{felzenszwalb2008discriminatively,lampert2009learning,farhadi2009describing,yu2010attribute,teo2015gestaltist,teo2013embedding,yu2011active} and activities are recognized by detecting the motions, objects and contexts involved in the activities \cite{laptev2005space,messing2009activity,wang2011action,gupta2007objects,conf/eccv/OgaleKA06,yang2014cognitive}.

Recently, researchers have advanced the viewpoint that if we are able to develop a semantic understanding of a visual scene, then we should be able to produce natural language descriptions of such semantics. This has given rise to a new area in the field that integrates vision, knowledge and natural language. Knowledge becomes especially important, as without background knowledge, it has become increasingly hard to obtain a desirable level of accuracy in this problem. And as such knowledge can be often mined from text, the problem now stands at the intersection between Computer Vision and Natural Language Processing. \textit{Mining such knowledge, storing it in a form that retains the semantics, and reasoning using this knowledge to develop a better understanding of scenes are the fundamental issues that are addressed in this paper.}

\begin{figure}[!h]
    \centering
    \subfigure[]{\includegraphics[height=0.09\textheight]{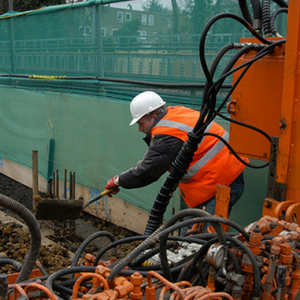} \label{fig:stanford_worker}}
    \subfigure[]{\includegraphics[height=0.08\textheight]{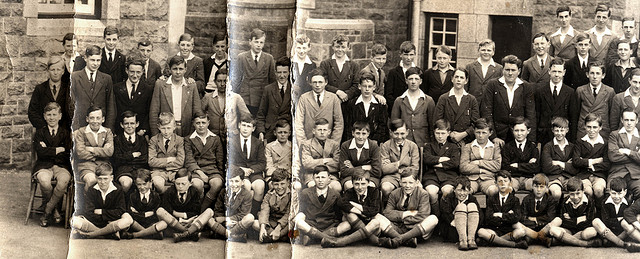} \label{fig:stanford_bananas}}
    \vspace*{-0.1in}
    \caption{Examples from \cite{karpathy2014deep}: (a) Positive example annotation: construction worker in orange safety vest is working on road, (b) Negative example annotation: a bunch of bananas are hanging from a ceiling. Such annotations could be infrequent, but it is hard to logically justify such contrasting outputs.}
    \label{fig:examples_stanford}
    \vspace*{-0.2in}
    \end{figure}

Current developments \cite{mao2014explain,kiros2014unifying,donahue2014long,karpathy2014deep,vinyals2014show,chen2014learning} in Computer Vision have shown that deep neural nets can be trained to generate a caption for an arbitrary scene with decent success. It is indeed an exciting achievement. However, current state-of-the-art image captioning systems still have a few drawbacks such as: 1) a brute-force image-to-text mapping makes it inconvenient to conduct Logical Reasoning beyond just doing inferences from annotated data; 2) due to the lack of intermediate semantic representations, they are all language-dependent; and 3) most importantly, when the system produces wrong results, it is almost impossible to trace back the system and analyze the failure case (See Figure \ref{fig:examples_stanford}). %This stems from the fact that current systems lack common-sense reasoning capability.

Let us consider how humans accomplish this task. Human perception is active, selective and exploratory. We continuously shift our gaze to different locations in the scene. After recognizing objects, we fixate again at a new location, and so on. We interpret visual input by using our knowledge of activities, events and objects. When we analyze a visual scene, visual processes continuously interact with our high-level knowledge, some of which is represented in the form of language. In some sense, perception and language are engaged in an interaction, as they exchange information that leads to meaning and understanding. Thus, our problem requires at least two modules for its solution: (a) a vision module and (b) a reasoning module that are interacting with each other. In this paper we propose to model the early stages of this process. The available datasets make it impossible to perform experiments that will consider vision as an active process, although this is the ultimate goal. Thus, our question becomes: if the vision module produces a number of (probabilistic) detections, how much the reasoning module can infer about the scene if it possesses common sense abilities? It turns out that the reasoning module can infer a great deal. %Let us give an example.  

  Motivated by such intuitions, we present here an effort to integrate deep learning based vision and state-of-the-art concept modeling from commonsense knowledge obtained from text. We use a deep learning-based perception system to obtain the objects, scenes and constituents with probabilistic weights from an input image. To predict how the objects interact in the scene, we build a common-sense knowledge base from image annotations along with a Bayesian Network capturing dependencies among commonly occurring objects and ``Abstract Visual Concepts'' (defined later). These two precomputed resources help us infer the following: 1) the correct set of correlated objects based on the high-confidence objects detected; 2) the most probable events that these objects participate in; 3) the role that the objects play in this event; and 4) given the events, objects and constituents, the ``Concept'' that emerges from such information.
 Based on these inferences, we output a Scene Description Graph (SDG) that depicts how these different entities and events interact.
 %\begin{figure}[!h]
 %\centering
 %\subfigure[]{\includegraphics[height=0.1\textheight]{figures/basketball1.jpg} \label{fig:basketball1}}
 %%\subfigure[]{\includegraphics[height=0.12\textheight]{figures/classroom1.jpg} \label{fig:classroom1}}
 %\caption{Two men playing basketball in an indoor stadium} % (b)few young students in a computer training classroom.}
 %\label{fig:examples}
 %\vspace{-0.1in}
 %\end{figure} 
 \begin{figure*}[htpb]
     \centering
     \subfigure{\includegraphics[height=0.1\textheight]{} \label{fig:basketball1}}
     \subfigure{\includegraphics[width=0.4\textwidth,height=0.09\textheight]{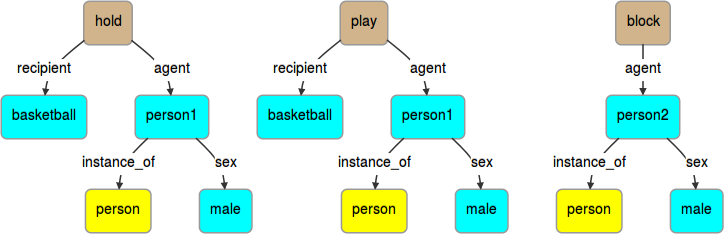}} % \label{fig:ideal1}}
     \subfigure{\includegraphics[width=0.4\textwidth,height=0.09\textheight]{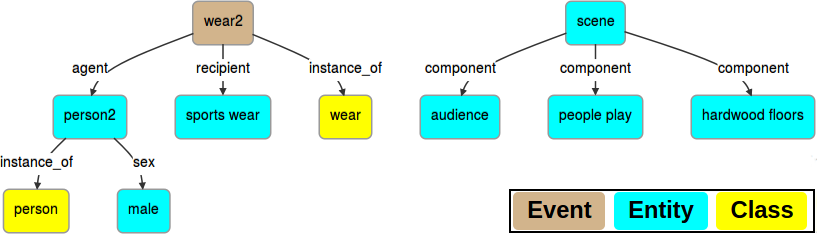}}% \label{fig:ideal2}}
     \vspace*{-0.1in}
     \caption{Example Image and a possible corresponding SDG. Note, the SDG should contain a similar event \textit{wear2} for \textit{person2}. We omit it for space constraints. Note that, it is easy to augment spatial information to the above graph such as (person1,left,person2).}
     \label{fig:example_knowledge_structure}
     \vspace{-0.15in}
     \end{figure*}
 In Figure \ref{fig:example_knowledge_structure}, we show a possible SDG for an example image. SDG is essentially a directed labeled graph among entities and events\footnote{Throughout this paper, we follow definition of Entities and Events from \cite{Gunning_projecthalo,Chaudhri:2007:EEB:1298406.1298435}.} that enables an array of possibilities to do further analysis beyond visual appearance, such as Event-Entity based analysis, question answering about the scene and flexible caption generation\footnote{One may note that such structures are also generated by Semantic parsers such as K-parser(\url{www.kparser.org}).}. %We argue that full vision, the capability to see and reason, should include not only deep appearance understanding of the scene, but also a deep common-sense reasoning at the concept level. We think that our approach paves a way towards solving the problems faced by current Scene Analysis systems by integrating vision with common sense reasoning. 
 
 The fundamental contribution of this work is a novel algorithm that uses automatically constructed Knowledge Base to create an SDG from an image, which facilitates further reasoning and caption generation. SDGs have advantages over ground-truth sentences because : 1) they can be easily processed by machines/AI systems in comparison to sentences; 2) the output can be rich in information-content; 3) they are not bounded by specific templates, that are  often  used by researchers to convert labels into sentences and 4) the SDG also can be used to generate sentence descriptions. 
 %To achieve this, we employ Deep Learning to perform object, scene and constituent recognition with state-of-the-art performance. 
 We also create a Knowledge Base which captures the knowledge about  the commonly-occurring Concepts, events and entities. %In our experiments, we capture events and entities pertaining to scenes describing images of people or dogs in natural setting. 
 The knowledge base can be used to provide answers to the following queries: 1) the event or set of events that connect two entities; 2) the role an entity plays in an event and 3) a subset of all possible concepts involving the entities and connecting events. Lastly, further inferences about the scenes such as ``Will the player holding the ball be able to tackle the blocker and under what conditions'' can also be attempted by feeding the SDG output as predicates to Reasoning modules along with additional background knowledge.

\vspace{-0.1in}
\section{Related Works}
\vspace{-0.05in}
Our work is influenced by various lines of work where researchers have proposed approaches to extract meaningful information from images and videos. As \cite{karpathy2014deep} suggests, such works can be categorized into 1) dense image annotations, 2) generating textual descriptions, 3) grounding natural language in images and 4) neural networks in visual and language domains.

According to the above categorization, we share our roots with the works of generating textual descriptions. This includes the works that retrieves and ranks sentences from training sets given an image such as \cite{hodosh2013framing}, \cite{FarhadiPictureStory},\cite{im2textOrdonez}, \cite{DBLP:journals/tacl/SocherKLMN14}. \cite{DBLP:conf/emnlp/ElliottK13}, \cite{Kulkarni11babytalk:}, \cite{Kuznetsova:2012:CGN:2390524.2390575}, \cite{Yang:2011:CSG:2145432.2145484}, \cite{journals/pieee/YaoYLLZ10} are some of the works that have generated descriptions by stitching together annotations or applying templates on detected image content.
   
%However, our aim is to reason beyond appearances. Though, we start with object, scene and constituents labels, we want to detect the concepts, more constituents and actions that occur in the scenes; which cannot be easily detected using even current state-of-the-art object or scene recognition techniques.

%The small number of works in Computer Vision, that aim to reason beyond appearance models, are also related to this paper. \cite{xie2013inferring} proposed that beyond state-of-the-art computer vision techniques, we could possibly infer implicit informations (such as functional objects) from videos, and they call them ``Dark Matter'' and ``Dark Energy''. \cite{joo2014visual} used a ranking SVM to predict the persuasive intention of the photographer who captured an image. More recently, \cite{pirsiavash2014inferring} seeks to infer the motivation of the person inside the image by mining knowledge stored in a large corpus using natural language processing techniques. These are fairly general investigations about reasoning beyond appearance.

Several works have shown %works, in which there have been 
promising efforts to acquire and apply commonsense in different aspects of Scene Analysis. \cite{conf/cvpr/ZitnickP13} uses abstraction to discover semantically similar images. \cite{DBLP:conf/cvpr/DivvalaFG14} proposes to learn all variations pertaining to all concepts and \cite{Santofimia12CommonSenseVision} uses common-sense to learn actions.

Recently, \cite{johnsoncvpr2015} introduced scene graphs to describe scenes and \cite{schuster-EtAl:2015:VL} creates scene graphs from descriptions. However, we automatically construct the graph from an image, and we believe, due to the event-entity-attribute based representation and meaningful edge-labels (borrowed from KM-ontology\cite{Clark2004}) , SDGs are more equipped to facilitate symbolic-level reasoning.
  %The rest of the paper is organized as follows. We briefly describe our approach to detect scenes, objects and constituents using the current state-of-the-art detection methods in Sec 3. In Sec 4, we provide small motivation for predicting Knowledge Structure and inferences from thought experiments introduced in the Introduction Section. Sec 5 describes our approach in inferring Knowledge Structure from images in detail. Finally Sec 6 and 7 describes the experiments, results and conclusion, future work respectively.
%\section{Our Approach}

\section{State-of-the-art Visual Detections}
\label{visualPrediction}
\vspace{-0.05in}
The recent development of deep neural networks based approaches revolutionized visual recognition research. Different from the traditional hand-crafted features, a multi-layer neural network architecture efficiently captures sophisticated hierarchies
describing the raw data \cite{Bengio2012}, which has shown superior performance on standard scene recognition \cite{zhou2014places}, object recognition \cite{girshick14CVPR} and image captioning \cite{karpathy2014deep} benchmarks.
%In the following sections, we first introduce the image dataset used for testing followed by details of each perception component.

{\bf Image Dataset: }In this paper, we use three image data sets, which are popularly referred to as Flickr 8k, Flickr 30k and Coco datasets \cite{hodosh2013framing}. These three datasets have 8,092, 31,783 and more than 160K images respectively. All the images from these datasets are accompanied with 5 hand-annotated sentences that describe the image. For all datasets, we used the train-test splits from \cite{karpathy2014deep} and the 4000 testing images (1000 each from Flickr 8k and 30k; 2000 from MS-COCO validation set; denoted as $I$) serve as the testing bed for our reasoning experiments. 

{\bf Deep Object Recognition: }We use the trained bottom-up region proposals and convolutional neural networks(CNN) object detection method from \cite{girshick14CVPR}. It considers 200 common everyday object classes (denoted as $\mathcal{N}$) and trained on ILSVRC 2013 dataset. We apply the method on the testing images($I$) and then convert the object detection scores to $P_r(n|I)$.

{\bf Deep Scene Recognition: }We use the trained CNN scene classification method from \cite{zhou2014places}. The classification model is trained on 205 scene categories (denoted as $\mathcal{S}$) and each of the category has more than 5000 training samples. We apply the method on the testing images and then convert the scene classification scores to $P_r(s|I)$.

{\bf Constituent Annotation Collection and Deep Constituent Recognition: } 
Images from the wild cannot always be categorized into a limited number of Scene categories. However, \textit{scene constituents} describing properties or actions of objects, attributes of scenes occur frequently across images and can be utilized to describe the image. %When we ask people to describe a scene, they tend to use more general phrases rather than just objects and scene, a.k.a. the constituents. 
In this work, we further augment the Flickr 8K image dataset with human annotation of constituents using Amazon Mechanical Turks. We specifically ask the human labeler to annotate not only objects, but what objects are doing or properties of objects. %Also we asked them to think about the general scenarios from image.

 %{\color{red} Not using clustering. Somak will update this paragraph}
We allow the labelers to use free-form text for describing constituents to reduce annotation effort. To obtain a standardized set of constituents from the annotations, we perform stop-words removal, parts-of-speech processing to retain nouns, adjectives and verbs. We replace the nouns with their superclasses such as \textit{man, boy, father} by \textit{person}, and then, we rank the resulting phrases according to their frequencies. Some of the top phrases are \textit{grass, dog run, dog play, kid play, person wear short}\footnote{All the phrases with their corresponding frequencies will be made publicly available in the final version.} etc. %We show the top $500$ phrases in the word-cloud in Figure \ref{fig:word-cloud}.

For the rest of our processing, we post-process the annotations for each training image and consider them if they are among the 1000 top constituents (denoted as $\mathcal{C}$). %Thus recognizing constituents on testing images becomes a multilabel classification problem.
Recent empirical results from a diverse range of visual recognition tasks indicate that the generic descriptors extracted from the CNN are very powerful \cite{donahue2014decaf,razavian2014cnn}. In this work, we use a pre-trained CNN from \cite{Krizhevsky2012}. %The convolutional layers conatin 96 to 1024 kernels of size 3x3 to 7x7. Max pooling kernels of size 3x3 and 5x5 are used at different layers to build  robustness to intra-class deformations. 
For each image in $I$, we use this pre-trained model to extract a $4096$ dimensional feature vector using \cite{donahue2014decaf,Krizhevsky2012}. We then trained a multi-label SVM to do constituents recognition using the deep features. The trained model is applied on all the testing images and we convert the classification scores to $P_r(c|I)$.

%{\bf Deep Image Captioning: }For evaluations, we use the implementation from \cite{karpathy2014deep} to generate a textual caption $S$ for each testing image. The method is based on a combination of CNN over image regions, bidirectional recurrent neural networks over sentences, and a structured multimodal embedding. We denote the set of captions as $S_I$.

%All of the visual modules used are using current tate-of-the-art deep neural network based method. 
The set of $P_r(n|I), P_r(s|I), P_r(c|I)$ makes up the initial visual perception output. % and we use $S_I$ for comparison and evaluation purposes. 

%\subsection{Reason beyond visual detections}
% using Text based Knowledge
\section{Constructing SDGs from Noisy Visual Detections}
    
\begin{figure}[t]
\begin{center}
\includegraphics[width=0.44\textwidth,height=0.4\textheight]{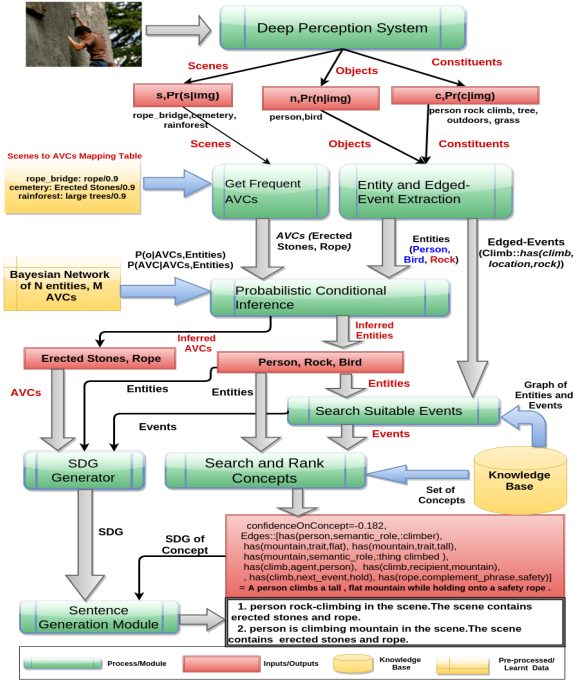}
\end{center}
\vspace{-10pt}
\caption{\small{SDG and Sentence Generation through Reasoning using Knowledge Base and a Bayesian Network $\mathcal{B}_{n}$}}
\label{fig:reasoning}
\vspace{-0.3in}
\end{figure}

    Next, we explain the reasoning framework to construct SDGs from noisy labels with the aid of knowledge from text. To provide a better understanding of this complex system, we provide a diagram of the architecture explaining the reasoning process for an example image in Figure \ref{fig:reasoning}.
    
    As shown, for each image, the above perception system produces  object, scene and constituent detection tuples. Each detection is provided with a confidence score. For objects, scores are provided for each bounding box. Top five scene labels and top ten constituent detections are considered for the reasoning framework. % in the following form:1) \textbf{Object Detection Tuples}: a collection of tuples such as \textit{\{basketball, $[B_1, score_1],[ B_2, score_2]$\}}, where $B_1$ and $B_2$ are two bounding boxes and $score_1$ and $score_2$ are confidence scores for each corresponding bounding box; 2) \textbf{Scene Detection tuples}: top five scene labels with confidence scores and 3) \textbf{Constituent Detection Tuples}: top ten  scene constituents with confidence scores.
    Most of these detections are quite noisy. We develop an elaborate reasoning framework to construct SDGs from such noisy detections, with the help of pre-processed background knowledge. %First we describe the data accumulation in the pre-processing phase and then describe our reasoning framework in detail.%To produce an SDG from such detections, we implement the following steps. (All the steps and boldened terms will be explained in the later subsections.)
\vspace{-0.05in}
\subsection{Pre-processing Phase Data Accumulation}
\vspace{-0.05in}
In this phase,  we collect Ontological information about object classes in Object Meta-data table ($\mathcal{O}_T$) and Scene classes in Scene Metadata ($\mathcal{S}_M$). We also store scene detection tuples ($\mathcal{S}_T$) and human annotation of images ($\mathcal{A}_d$) for all training images.
 We create a \textbf{Knowledge Base} $\mathcal{K}_b$, a \textbf{Bayesian Network} $\mathcal{B}_{n}$ and a Scenes to Abstract Visual Concepts\footnote{\textit{Abstract Visual Concepts} are higher-level scene constituents and they describe commonly occurring visual concepts that can be observed across images. In essence, they are can be compared to phrasal verbs. Like textual phrases, they do not necessarily follow compositionality of its constituent words. In case of AVC, this happens in the context of images. For example \textit{waiting room} suggests room, seating area, chairs, people waiting etc. 
    
    In comparison, \textit{Scene Constituents} can be compared with phrases that retain their compositionality such as \textit{people play, person wear shorts} etc. In context of an image, \textit{person wear shorts} can be grounded as the objects \textit{person} and \textit{shorts}; and the action \textit{wear}.}  (AVC) Mapping Table ($\mathcal{S}_M$).
    
  %For the pre-processing phase, we accumulate metadata such as scene detection tuples, image annotations, object meta-data and scene meta-data.
  
  \textbf{Scene Detection tuples ($\mathcal{S}_T$):} We use the perception system of the previous section to create scene detection tuples ($\{(s_i,P_r(s_i|I_{tr}))| i \in \{1,..,5\}\}$) of set of training images ($I_{tr}$). These are used to learn the Bayes Net $\mathcal{B}_{n}$. 
  %The perception system of the previous section produces object, scene and constituent tuples. We use only scene detection tuples of training images to help build the table $\mathcal{P}_{JFT}$. 
  
  \textbf{Image Annotations ($\mathcal{A}_d$):} We collect all textual descriptions of the training images provided with Image Datasets and use them for building the knowledge base $\mathcal{K}_b$ and Bayes Net $\mathcal{B}_{n}$. However, both can be built using \textbf{any repository of sentences} that describe day-to-day concepts.
   
 \textbf{Object Meta-data ($\mathcal{O}_T$):} For each of the 200 object classes, we collected all synonyms, hyponyms and hypernyms. The list is prepared using Wordnet API. This is dataset-independent and only needs to be augmented when the set of object classifiers expands.
 
 \textbf{Scenes-to-AVCs Mapping Table ($\mathcal{S}_M$):} For each scene in $\mathcal{S}$, we added ontological information involving a set of abstract concepts and a set of synonyms. To obtain the synonyms, we again used WordNet API. We hand-annotated all the AVCs for each scene and learnt a prior belief for each AVC in scene from human annotations. For example, for the scene \textit{airport\_terminal}, we add \textit{\{waiting room, big glass view, people\}} as the list of AVCs and \textit{terminal} as the synonym; and learn the priors $0.7,0.6\text{ and }0.9$ respectively for AVCs. %given in Table \ref{table:scene_metainfo_table}.
    %At this point, we should clarify that, \textit{abstract concepts} are higher-level scene constituents and they describe commonly occurring visual concepts that can be observed across images.%. \textit{Scene constituents} are entities, properties of entities, actions involving entities occurring in a scene.
  
% \begin{table}[!htpb]
%       %\vspace{-10pt}
%        \centering       
%\resizebox{\linewidth}{!}{%
%        \begin{tabular}{|l|l|l|}           
%        \hline  {Scene Label} & {Abstract Concepts with prior} & {Synonyms}           
%        \\   
%        \hline   {airport\_terminal} & 
%        \begin{tabular}[x]{@{}l@{}}
%        waiting room(0.7), \\
%        big glass view(0.6) \\
%         people(0.9)
%        \end{tabular}
%        &  terminal 
%        \\
%        \hline
%        \end{tabular}
%}
%        \caption{This table provides meta-information for the scene label ``airport\_terminal''. We should clarify that, abstract concepts are higher-level scene constituents. Scene constituents are entities, properties of entities, actions involving entities occurring in a scene.}
%        \label{table:scene_metainfo_table}
%        \vspace{-0.1in}
%         \end{table}

In the following sub-sections, we first introduce the Reasoning Framework briefly, followed by a description of the construction of  the Knowledge Base $\mathcal{K}_b$ and the Bayesian Network $\mathcal{B}_{n}$. Lastly, we describe our reasoning framework in detail.
    
\vspace{-0.05in}
\subsection{Reasoning Framework}
\vspace{-0.05in}    
    Equipped with the background knowledge stored in the form of  ($\mathcal{K}_b,\mathcal{B}_{n},\mathcal{S}_M,\mathcal{O}_T$), we process the objects, scene and constituent detections for an image to construct an appropriate SDG in the following way: i)  we populate synonyms, hypernyms, hyponyms of objects and synonyms, AVCs (with priors) of scenes; ii) (\textbf{Scene Constituents}:) we extract entities and events from each constituent. Such as, the constituent \textit{person wear short} results in an event \textit{wear} with two edges: one labeled \textit{agent} joining the entity \textit{person} and another labeled \textit{recipient} joining the entity \textit{short}; iii) (\textbf{Abstract Visual Concepts}:) we choose the AVCs iteratively that maximizes the conditional probability given high confidence objects; iv) (\textbf{Objects}:) for low-scoring objects, we choose the sibling (in the hyponym-hypernym hierarchy) which maximizes the conditional probability given high-confidence objects and AVCs; v) (\textbf{Events}:) we search the $\mathcal{K}_b$ to find the most compatible events %(or \textit{chain of events}) 
    that connect pairs of high-confidence objects. We add the events obtained from \textbf{Constituents} to this set of compatible events; vi) (\textbf{Concepts \footnote{The idea behind the representation of a \textit{Concept} is inspired by that of a \textit{Process} in AURA \cite{Chaudhri:2007:EEB:1298406.1298435}. The structure \textit{Process} is a graph that represents a Biological Process in AURA-KB. and it symbolizes a higher-level event that encapsulates smaller events, and the entities that participate in such events. Similarly, a \textit{Concept} represents a natural activity where a few events occur and participating entities interact through the events.
 Our entire approach, in essence, is about sequentially determining the participants (entities and events) of a \textit{Concept} and lastly, the \textit{Concept} itself.}}:) given the above events and AVCs,  we search the $\mathcal{K}_b$ for Concepts that best suits the events and AVCs, and  we also construct an SDG  based on just high-confidence objects, events and AVCs.
 
  % \end{itemize}

\vspace{-0.03in}
\subsection{Knowledge Base ($\mathcal{K}_b$)}
\begin{figure}[t]
         \begin{center}
         \subfigure[]{\includegraphics[width=0.45\textwidth,height=0.25\textheight]{figures/image_perception.png}}
         \subfigure[]{\includegraphics[width=0.4\textwidth,height=0.14\textheight]{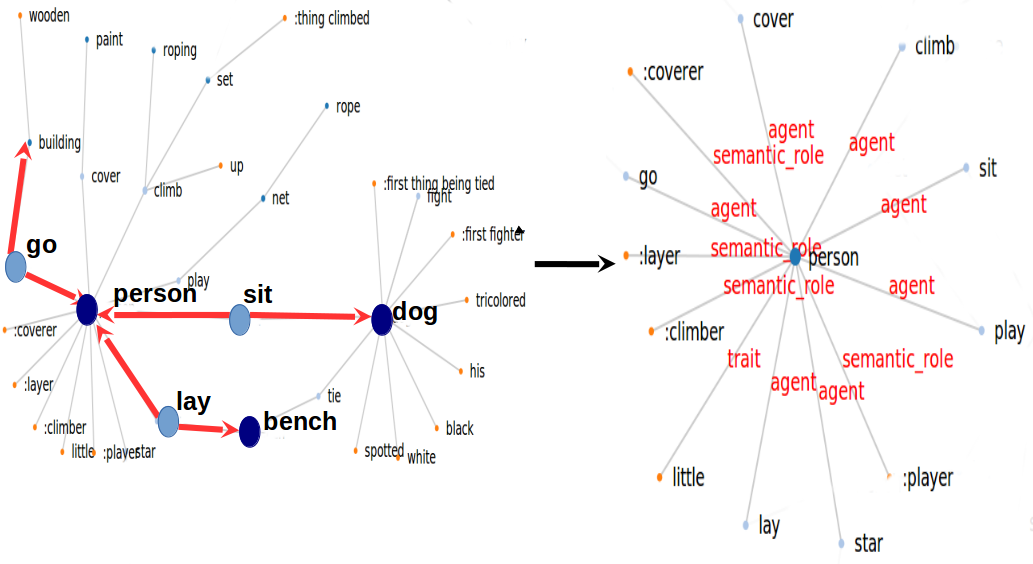}} 
         \end{center}
         \vspace{-13pt}
         \caption{\small{(a) Constructing Knowledge Base From Annotations. (b) A snapshot of the $\mathcal{K}_b$. In this figure, \textit{Person} and \textit{bench} are entities, \textit{lay} is the connecting event. The entity \textit{Person} can have trait \textit{climber}. The sub-graph essentially captures the knowledge of the activity \textit{person laying on a bench}. The figure on the left shows the edge-labels.}}
         \label{fig:kparser1}
         \vspace{-0.25in}
         \end{figure}

\vspace{-0.03in}
     %We use the K-Parser (available at www.kparser.org, \cite{DBLP:conf/ijcai/SharmaVAB15}), to parse each sentence from collection of annotated sentences, generalize and store the data as a knowledge base. The K-parser, in-turn, uses Stanford Dependency parsing to get a dependency graph and then maps these dependencies to meaningful labels according to KM-ontology. In Figure \ref{fig:kparser1}, we describe how we construct our knowledge base using the Stanford Parser and K-Parser.
In Figure \ref{fig:kparser1}, we describe how we construct $\mathcal{K}_b$ from a set of Image Annotations ($\mathcal{A}_d$) using the Stanford Parser and K-Parser \cite{DBLP:conf/ijcai/SharmaVAB15}. For each sentence, we first parse using the Stanford Parser to get a dependency graph. The K-parser then maps these dependency labels using a set of rules to a set of meaningful labels from KM-Ontology\cite{Clark2004} and the resulting graph is further augmented using ontological and semantic information from different sources (more details on \url{kparser.org}). We then generalize each of these graphs i.e. replace entities by their superclasses. Then  we merge them based on overlapping entities and events, and create a single graph ($\mathcal{K}_b$).
     %One example of parsing by K-parser is given in Figure \ref{fig:kparser1}.
       
$\mathcal{K}_b$ is defined as the tuple $(\mathcal{G},\mathcal{C})$. $\mathcal{G}=(V,E)$ denoting set of vertices $V$, set of edges $E$.  Each vertex and edge has a label. Each vertex can be of three types: \textit{events, entities} and \textit{traits}. \textbf{Events} correspond to verbs, \textbf{entities} correspond to superclasses of nouns that directly interact with events and \textbf{traits} represent all other nouns.  \textbf{Edge labels} in the $\mathcal{K}_b$ are exactly the same as in the K-parser (Figure \ref{fig:kparser1}). $\mathcal{C}$ is a set of \textbf{concepts} which corresponds to generalized K-parser graphs of sentences and is essentially a sub-graph of $\mathcal{G}$.

   From Flickr8k annotations, we process nearly $16000$ sentences provided for the images ($2$ per each image\footnote{We do not consider all 5 sentences for an image, as most of the sentences are conceptually same.}) to build the $\mathcal{K}_b$. %For each sentence, the K-Parser graph is generalized using superclass relations and merged into a single graph of entities and events.  
   A visualization of a part of the $\mathcal{K}_b$ is given in Figure \ref{fig:kparser1}(b).
   %A summary of the statistics of the Knowledge-Base is provided in the following table. {\color{red} Somak, create KB stats table.}
   After parsing the annotated sentences in Flickr8k, $\mathcal{K}_b$ consists of $1102$ events, $2500$ entities and $1869$ traits. The total number of edges and distinct concepts in the graph are $25271$ and $14325$.

%   \begin{figure}[t]
%    
%\begin{center}
%\includegraphics[width=0.48\textwidth,height=0.13\textheight]{figures/kb_modified_edges.png}
%\end{center}
%\vspace{-13pt}
%\caption{\small{ A snapshot of the KnowledgeBase $\mathcal{K}_b$. In this figure, \textit{Person} and \textit{bench} are entities, \textit{lay} is an event that connect such entities. The entity \textit{Person} can have trait \textit{climber}. Some of the nodes and edges are boldened to clearly represent the directions and labels on nodes and edges. }}
%\label{fig:kb1}
%\vspace{-0.2in}
%
%\end{figure}

\vspace{-0.08in}
% using Frequency Table $\mathcal{P}_{JFT}$
\subsection{Conditional Probability Estimation}
\vspace{-0.05in}
  In this sub-section, we describe the type of conditional probabilities we estimate and the Bayesian Network we learn to estimate such probabilities.
  We use conditional probability calculations in two of the steps of our approach: inferring the most probable collection of Abstract Visual Concepts and rectifying low-scoring erroneous objects.
    
    For inferring the most probable collection of AVCs, we first make a list $(C_{freq})$ of all the frequent AVCs (with frequency $> 2$ in our experiments) from all scenes detected  for a test image. Then we follow Algorithm \ref{algo:inferConcepts} to get the set of inferred concepts $C_{inf}$ from the set of high-scoring (score $>\alpha_h$\footnote{The hyper-parameter ($\alpha_h$) are set based on performance on validation data.}) entities $\mathcal{O}_{img}$ and the set of scenes $\mathcal{S}_{img}$ detected for image $img \in I$. We iterate till the entropy keeps decreasing.
\vspace{-0.15in}
\begin{algorithm}
\caption{Infer Abstract-Visual-Concepts}\label{algo:inferConcepts}
{\fontsize{7}{7}\selectfont
\begin{algorithmic}[1]
\Procedure{inferMostProbableAVCs}{$\mathcal{S}_{img}, \mathcal{O}_{img}, C_{freq}$}
\State $prevH \gets 1$
\While {$C_{freq} \neq \phi$} 
\State $
S_{max} \gets \arg\!\max_{s \in C_{freq}} P(s | C_{inf}, \mathcal{O}_{img})$
\State$
H \gets \sum_{s \in C_{freq}} \{-P(s | C_{inf}, \mathcal{O}_{img}) * log P(s | C_{inf}, \mathcal{O}_{img})\}
$
\If{$H > prevH$} break; \EndIf
\State $prevH \gets H$
\State $C_{inf} \gets C_{inf} \cup \{S_{max}\}$; $C_{freq} \gets C_{freq}\setminus \{S_{max}\}$
%\State $C_{freq} \gets C_{freq}\setminus \{S_{max}\}$
\EndWhile
\EndProcedure
\end{algorithmic}
}
\end{algorithm} 
\vspace{-0.15in}

  Next, we attempt to rectify the low-scoring entities based on high-scoring entities ($\mathcal{O}_{img}$) and the above $C_{inf}$. For each low-scoring entity, we get all its siblings i.e. we get all the children of its hypernyms. For example, if \textit{bathing cap} is assigned a low score, the assigned superclass is \textit{headwear} and its children are \textit{headband, hat} etc. We calculate the following $o_{max} = \arg\!\max_{o \in siblings} P(o | C_{inf}, \mathcal{O}_{img})$ and then add $o_{max}$ to the high-scoring entities list ($\mathcal{O}_{img}$).
  
  As the above paragraphs suggest, we need to estimate the conditional probabilities:  $P(s | C_{inf},\mathcal{O}_{img})$ and $P(o | C_{inf}, \mathcal{O}_{img})$. To estimate the conditional probabilities, we learn a Bayesian Network $\mathcal{B}_{n}$ using $\mathcal{A}_d$ and $\mathcal{S}_T$. 
 
\vspace{-0.15in}
\subsubsection{Learning the Bayesian Network $\mathcal{B}_{n}$}
\begin{figure}[!hpt]
 \begin{center}
 \includegraphics[width=0.44\textwidth,height=0.1\textheight]{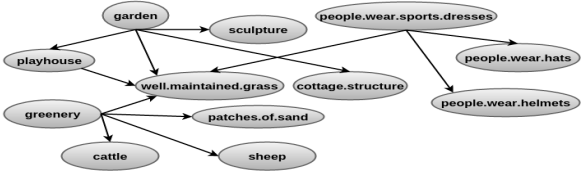}
 \end{center}
 \vspace{-10pt}
 \caption{\small{A subgraph reflecting the dependencies captured in the Learnt Bayesian Network $\mathcal{B}_{n}$}}
 \label{fig:bn}
 \vspace{-0.15in}
 \end{figure}

\vspace{-0.1in}
  To capture the knowledge of naturally co-occurring entities and Abstract Visual Concepts, we learn a Bayesian Network that represents the dependencies among them.
    We create the training data $\mathcal{D}$ which is a set of tuples $T=(t_i)_i, i \in {1,..,N}$ where $N$ is the total number of entities and AVCs. Each term $t_i$ is binary and denotes 1 if the $i^{th}$ entity (or AVC) occurs in the tuple. Then, we use the Tabu Search (tabu) algorithm  to learn the structure and then we populate the Conditional Probability Tables using the R-bnlearn package \cite{bnlearn}. A subgraph of the learnt Bayesian Network is shown in the Figure \ref{fig:bn}.
 
    To create the training data $\mathcal{D}$,  we process each training image (in $I_{tr}$), and we automatically detect entities and AVCs and then output the tuple $T$. To detect entities, we parse the image annotations ($\mathcal{A}_d$) and extract entities from it. Some of the AVCs such as \textit{people and people wear shorts} are detected using rule-based techniques. However, for scenes such as \textit{airport-terminal}, it is unlikely that AVCs such as \textit{waiting room} can be found in human descriptions of an image; as we tend to describe only the entities and their interactions. Keeping this idea in mind, we ran the scene classifier system from the previous Section \ref{visualPrediction}, and we consider all the AVCs of the scene with the highest score ($Pr(s|I_{tr})$), from the \textit{Scene-to-AVC lookup table} ($\mathcal{S}_M$).
\vspace{-0.05in}	
\subsection{Ranking and Inferring Final Concepts}
\vspace{-0.05in}
	Given the most relevant set of Abstract Visual Concepts ($\mathcal{C}_{inf}$) and entities ($\mathcal{O}_{img}$), we find Concepts that the image describes. To do this, we use the $\mathcal{K}_b$ to search first for events that these entities (i.e. objects) participate in and then we use these events and entities together to search for Concepts in the set of concepts $\mathcal{C}$ in $\mathcal{K}_b$.
	
	We rely on two assumptions about the Knowledge Base: I) $\mathcal{K}_b$ reflects a more-or-less complete view of the relevant world knowledge and hence we can find the most suitable events from it. This assumption is valid if the images come from the same domain; such as in our examples, we have used the Flickr8k dataset and the domain corresponds to pictures of humans and dogs in natural setting; and II) $\mathcal{K}_b$ contains all concepts possible with the given events and entities. 
	This is a strict assumption, which might not be true even if we parse the whole Web. To alleviate the problem, we give two final outputs: i) an SDG involving the entities, AVCs and events and ii) another SDG of the top Concept that is obtained from $\mathcal{C}$ in $\mathcal{K}_b$.
	%The rest of our approach is described as follows.

 \textbf{Search Connecting Events}: 
 The motivation behind building a Knowledge Base was to logically explain why certain co-occurring events are suitable for the combination of entities. For example, consider the entities \textit{person} and \textit{swimming trunks}. Note, \textit{swimming trunks} corresponds to the vertex \textit{trunk} in $\mathcal{K}_b$. We get events such as sniff, climb, wear etc., i.e., some corresponding to tree-trunk and others to swimming-trunks. To logically find suitable events, we find all connecting events from $\mathcal{G}$ in $\mathcal{K}_b$ and then filter spurious events based on ontological and background knowledge from $\mathcal{O}_T$ and $\mathcal{C}$ in $\mathcal{K}_b$.

 For a pair of entity in $\mathcal{O}_{img}$, we traverse the path from one entity to another in the graph $\mathcal{G}$ and consider event-nodes on the path.
 As shown in Figure \ref{fig:kparser1}(b), two entities can be connected by an event. However, in some cases, they could be connected by a chain of events and entities. We employ a greedy breadth-first search over the graph $\mathcal{G}$ for such pairs. We denote the set of entities that are related to each other by some event, by $\mathcal{O}_{ev}$.
% 
% The motivation behind building a Knowledge Base was to logically explain why certain co-occurring events are suitable for the combination of entities. For example, consider the entities \textit{person} and \textit{swimming trunks}. Note, \textit{swimming trunks} corresponds to the vertex \textit{trunk} in $\mathcal{K}_b$. We get events such as sniff, climb, wear etc., i.e., some corresponding to tree-trunk and others to swimming-trunks.
 
 For filtering spurious events, we introduce the notion of \textit{Edge-Compatible Events}.
% 
%  %We define \textbf{edge-compatibility} of an event for two entities as follows: 
  An event is \textbf{edge-compatible} with respect to two entities if they are connected to the event using edges with compatible labels. 
  As these labels are well-defined relations between entities and events from KM-Ontology, %As these labels are from KM-ontology, they preserve the semantics. So, 
  the label-compatibility is easy to observe. For example, \textit{(agent,recipient)} is a compatible pair and only an animate entity can be an \textit{agent}. Based on the rules, the event \textit{wear} is edge-compatible with respect to entities \textit{person} and \textit{trunk}.

 Even after this, we still obtain events like \textit{climb} etc. To filter such events, we consult the table $\mathcal{O}_T$ and the set of concepts $\mathcal{C}$. We know that the entity \textit{swimming trunks} belongs to the superclass \textit{clothing}, and hence we retain only those events that are connected to an entity \textit{trunk} which is of the same superclass, in some concept in $\mathcal{C}$.
    
  \textbf{SDG Construction:} After obtaining a set of suitable events (such as \textit{wear}), we construct an SDG using the following set of rules: i) add \textit{has(scene, component, $s$)} for all AVC 
 $s$ in $C_{inf}$; ii) add \textit{has(event, location, scene)} for the top detected events; iii) add all compatible edges related to the events such as \textit{has(wear,agent,person)} and  \textit{has(wear,recipient,trunk)}; and iv) for all entities $o_{im}$ in $(\mathcal{O}_{img}\setminus\mathcal{O}_{ev})$, do the following: if it is an animate entity, add \textit{has($o_{im}$, location, scene)}; Otherwise, find the shortest path from $o_I$ to the top detected event in the $\mathcal{K}_b$ and add the edges on the path to the SDG.
 %\end{itemize}
 
 \textbf{Search Concepts}: Given the events and entities ($\mathcal{O}_{ev}$), we search the set of Concepts $\mathcal{C}$ in $\mathcal{K}_b$. % and retain those which has all the labeled edges corresponding to the events.
 Recall, in the $\mathcal{K}_b$, a Concept is a generalized K-parser graph of a sentence. We consider a Concept as candidate if all edges from a detected Edge-Compatible Event are present in it.
 %For each such concept, we search all its edges and we retain the concept if and only if all edges from the event are present in it.
 
 Next, we weight each candidate Concept using the remaining entities in $(\mathcal{O}_{img}\setminus\mathcal{O}_{ev})$ and AVCs; i.e., increase a counter if an entity or AVC occurs in the graph. We also calculate a joint confidence-score for each Concept based on the $P_r(n|I), P_r(s|I), P_r(c|I)$ values of the object, scene and constituents present in the Concept. Based on the counters and the joint confidence-score, we rank the Concepts.

 %\item \textbf{Superclass}: We also filter the events
 
  \textbf{Template Based Sentence Generation}: We  generate textual descriptions from the SDG using the SimpleNLG\cite{Gatt:2009:SRE:1610195.1610208} package.
  For example, for the edges \textit{has(wear,agent,person)} and \textit{has(wear,recipient,shorts)}, we will generate the sentence ``\textit{a person is wearing shorts}''. Based on the edge-labels (labels from KM-ontology) we populate the verb, subject, object and adjectives (including quantitative\footnote{For high-scoring detections, we also consider the spatial information from the bounding-boxes. For $N$ such detections of an object $obj$, we generate sentences like \textit{$N$ $obj$'s are in the scene}. }) of sentences using simple rules. %The word \textit{trunk} is replaced by \textit{swimming trunks} by consulting the concepts in knowledge-base $\mathcal{K}_b$.
   It should be noted that these K-Parser labels are a direct mapping from the set of Stanford Dependencies, and theoretically we can populate all the parts-of-speeches of a sentence from the SDG. Herein lies the effectiveness of producing an SDG from an image. %It is both consumable by symbolic reasoning procedures and easily convertible to textual descriptions which can facilitate human evaluations of the system.  

%\vspace{-0.1in}
\vspace{-0.1in}     
\begin{figure*}[!t]
     \centering
      \subfigure[]{\includegraphics[width=0.08\textwidth, height=0.09\textheight]{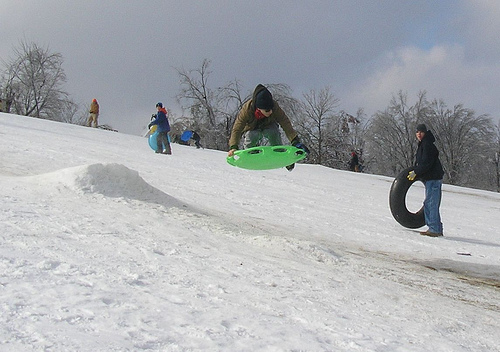} \label{fig:rich1}}
      \subfigure[]{\includegraphics[width=0.22\textwidth, height=0.09\textheight]{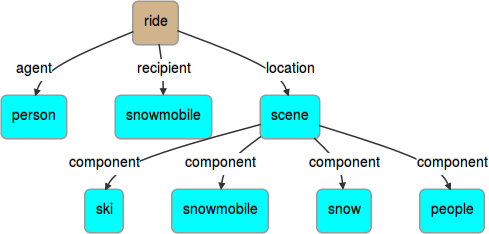} \label{fig:image1_rich1}}
      \subfigure[]{\includegraphics[width=0.07\textwidth, height=0.09\textheight]{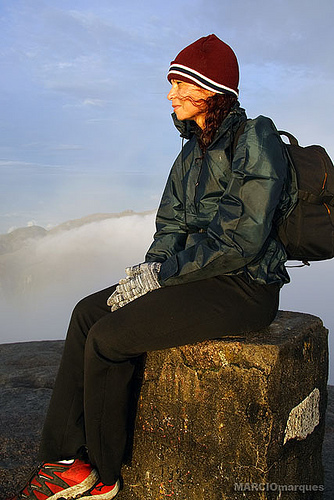} \label{fig:rich4}}
      \subfigure[]{\includegraphics[width=0.24\textwidth, height=0.09\textheight]{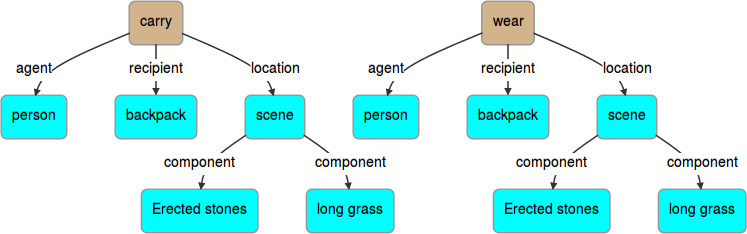} \label{fig:image4_rich4}}
      \subfigure[]{\includegraphics[width=0.07\textwidth, height=0.09\textheight]{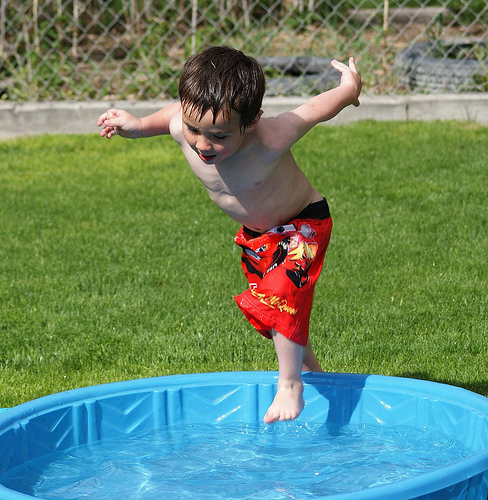}  \label{fig:rich3}}
     \subfigure[]{\includegraphics[width=0.25\textwidth, height=0.09\textheight]{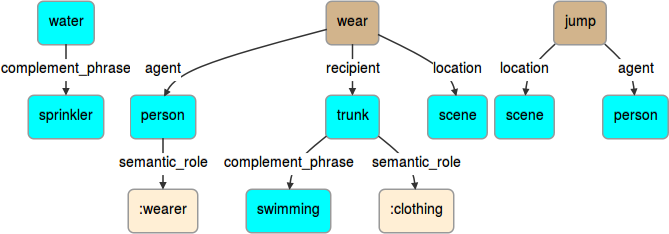}  \label{fig:image3_rich3}}
     
     \subfigure[]{\includegraphics[width=0.15\textwidth,height=0.09\textheight]{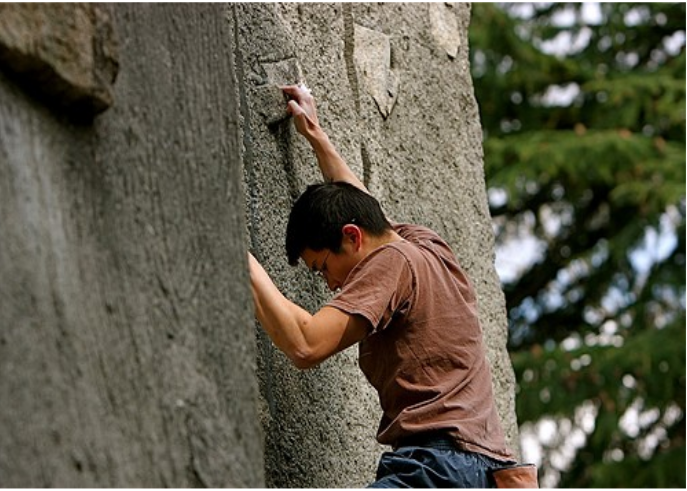} }
           \subfigure[]{\includegraphics[height=0.09\textheight]{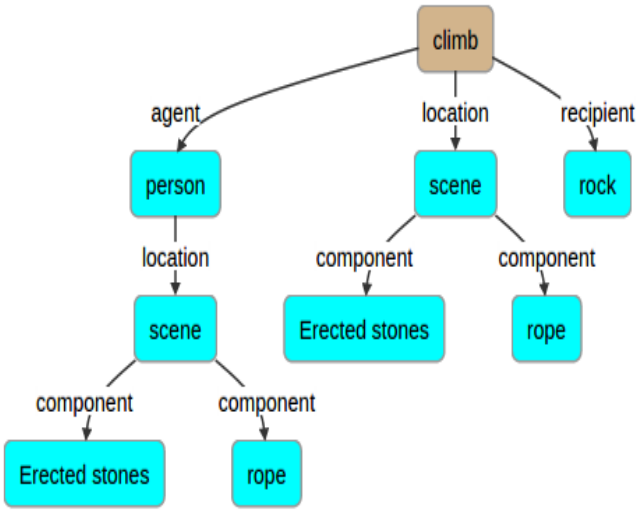} }
           \subfigure[]{\includegraphics[width=0.15\textwidth,height=0.09\textheight]{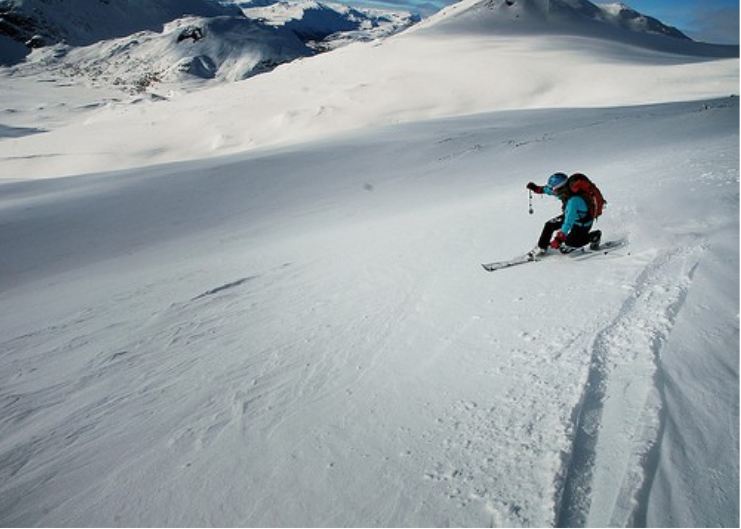}  }
          \subfigure[]{\includegraphics[width=0.50\textwidth,height=0.09\textheight]{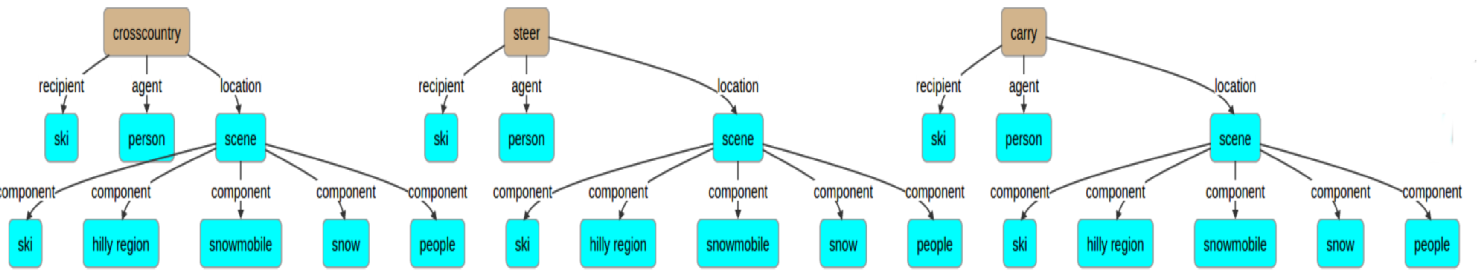}}
     %\subfigure[]{\includegraphics[scale=0.8]{fig/grasp/260.png} \label{fig:four_types_2}}
     \vspace{-0.15in}
     \caption{\small{The SDGs in (d), (e) and (f) corresponds to images (a), (b) and (c) respectively. \textbf{For more detailed examples, please check Appendix and \url{http://bit.ly/1NJycKO}.}}}
     %As you can observe, due to such homogeneous semantic representation, sentences and images both can be represented seamlessly in such graphical format. }
     \label{fig:graph_rich}
     \vspace*{-0.25in}
\end{figure*} 

\section{Experiments and Results}
\vspace{-0.05in}
 %Unlike standard caption generation tasks, the evaluation of the afore-mentioned Knowledge-Structure is not exactly straightforward.  
 The Knowledge-Structure representing a scene should be rich in information-content and should carry enough semantics to describe the image. %The best way for evaluating such a structure and in turn, such a system is by Question-answering. But,in the absence of ``annotated''(for QA) datasets elsewhere, we resorted to the following evaluation strategy for our system. 
 We adopted three sets of experiments. First, we detect the accuracy with which our system can detect events and entities present in the image. We perform a qualitative evaluation (``relevance'' and ``thoroughness'') of the textual descriptions generated from SDGs with the sentences generated by \cite{karpathy2014deep} using the Amazon Mechanical Turkers (AMT). And lastly, to evaluate the image-sentence alignment quality, we design an Image Retrieval task and report our results on Image Search based on generated annotations. To conclude, we provide a few example images and their SDGs. 
 
 For comparison purposes, we use the implementation from \cite{karpathy2014deep} to generate a textual caption $S$ for each testing image. The method is based on a combination of CNN over image regions, bidirectional recurrent neural networks over sentences, and a structured multimodal embedding. We denote the set of captions as $S_{NN}$.
  %A related paper \cite{johnsoncvpr2015} provides an evaluation using gold-standard of a few images. We found this method to be unreliable as such a Gold-standard for huge complicated graphs could vary among annotators. 
  
\textbf{Training Phase: } Our model can be represented by the tuple $(\mathcal{K}_b,\mathcal{B}_{n},\mathcal{S}_T,\mathcal{A}_d,\mathcal{O}_T,\mathcal{S}_M)$. Among these, $\mathcal{S}_T,\mathcal{O}_T \text{ and }\mathcal{S}_M$ are collected and stored once, and re-used for all datasets. For our experiments, we re-use the same Bayesian Network $\mathcal{B}_{n}$ learnt from Flickr8k data for all the datasets. Though, we build the $\mathcal{K}_b$ each time from the annotated sentences, this can be easily avoided by using the same $\mathcal{K}_b$ for all the datasets. In essence, for the reasoning part, we donot require any training at all for new datasets.

\textbf{Entity and Event Detection Accuracy: }
%\vspace{-0.05in}
 For this experiment, we extracted entities and events (gold-standard) from constituent annotations for the 1000 test images of Flickr8k. We manually checked them to remove noise. To provide a baseline, we also extracted entities and events from $S_{NN}$ automatically using K-parser. Subsequently, we compared the gold-standard with entity-event set from \cite{karpathy2014deep} and the SDG output from our system for each image. The statistics of our evaluation is given in Table \ref{table:stats_table}.
 %Note that the accuracies are comparable even though \cite{karpathy2014deep} output comes from the set of annotation sentences, and we are detecting entities, traits and events from scratch. 
    %One should note that, we have been strict in the above evaluations. While detecting expected events and entities, we have collected all annotated sentences (i.e. 5 different descriptions of same image). This adds up to quite a variety of events and entities.
\begin{table}[!htpb]
\vspace{-0.1in}
\centering
\resizebox{\columnwidth}{!}{%
%\normalsize {
\begin{tabular}{|l|l|l|l|l|}           
\hline  {Type} & {Accuracy-SDG($\%$)} & {Precision-SDG($\%$)}  & \begin{tabular}[x]{@{}c@{}}Accuracy($\%$)-\\\cite{karpathy2014deep}\end{tabular} & \begin{tabular}[x]{@{}c@{}}Precision($\%$)-\\\cite{karpathy2014deep}\end{tabular}       
\\   
\hline   Entities & $13.6$ & $21.7$ & $16.9$ & $34.2$
\\
\hline Events &$13.1$ & $8$ & $15.3$ & $15.2$
%\\
%\hline Edges &$3.6$ & $5$ & $1$ & $5$
\\
\hline
\end{tabular}
}
%}
\vspace{-0.1in}
\caption{Accuracy and Precision in events and entities prediction}
\label{table:stats_table}
\vspace{-0.15in}
\end{table}

  %Here we want to mention a caveat: the testing images which we think are semantically rich are manually selected, in order to better show how the human-like reasoning process can further rectify visual recognition and build a deep understanding of the image. With the current rapid development in visual recognition community, many of the them are approaching human-level performance.  We believe the same process we presented here can be further applied on general cases as well.

\textbf{AMT Evaluation of Generated Sentences: }
%\vspace{-0.05in}
  Since sentence generation to describe a scene is innately a creative process, a good metric is to ask humans to evaluate these sentences. The evaluation metrics: Relevance and Thoroughness, are therefore proposed as empirical measures of how much the description conveys the image content (relevance) and how much of the image content is conveyed by the description (thoroughness)\footnote{For complete instructions provided to the turkers, please check out Appendix.}. We engaged the services of AMT to judge the generated descriptions based on a discrete scale ranging from 1--5 (low relevance/thoroughness to high relevance/thoroughness). The average of the scores and their deviation are summarized in Table~\ref{tab:sengen_res1} for Flickr8k, Flickr30k test images and MS-COCO validation images. For comparison, we asked the AMTs to also judge one random gold-standard description and the output from \cite{karpathy2014deep}, a state-of-the-art image captioning system.
     
     In our experiments, we found that $\mathcal{K}_b$ from Flickr8k annotations can be used for Flickr30k without much effect on accuracy. However, for MS-COCO datasets, $\mathcal{K}_b$ from Flickr8k annotations falls short of producing a desired accuracy as the COCO data is much more varied.
%\textit{For Flickr 30k and MS-COCO, we have reused the $\mathcal{K}_b$ created from Flickr8k annotations. For Flickr30k, our approach  fares well as the images come from the same domain. Though MS-COCO data is more varied, we achieve a comparable relevance and thoroughness\footnote{The re-use of Flickr8k $\mathcal{K}_b$ and the associated performance proves how ``domain-related'' knowledge collected from any source can be applicable to any datasets from the same domain.}.} %We have used the same train-test-val splits as mentioned in \cite{karpathy2014deep}.}
%\vspace*{-0.1in} 
\begin{table}[!htbp]
   \begin{center}
   %{\footnotesize
   \resizebox{\columnwidth}{!}{%
   \begin{tabular}{|p{2.0cm}||c|c|c|}
   \hline 
   Experiment & \cite{karpathy2014deep} & Our Method & Gold Standard\\ 
   \hline 
   R $\pm$ D(8k) & $2.08 \pm 1.35$ & \textbf{2.82 $\pm$ 1.56} & $4.69 \pm 0.78$\\\hline
   T $\pm$ D(8k) & $ 2.24 \pm 1.33$ & \textbf{2.62 $\pm$ 1.42} & $4.32 \pm 0.99$\\\hline
   \hline
   R $\pm$ D(30k) & $1.93 \pm 1.32$ & \textbf{2.43 $\pm$ 1.42}  & $4.78 \pm 0.61$\\\hline
   T $\pm$ D(30k) & $ 2.17 \pm 1.34$ & \textbf{2.49 $\pm$ 1.42} & $4.52 \pm 0.93$\\\hline
    \hline
   R$\pm$D(COCO) & \textbf{2.69 $\pm$ 1.49} & $2.14 \pm 1.29$  & $4.71 \pm 0.67$\\\hline
   T$\pm$D(COCO) & \textbf{2.55 $\pm$ 1.41} & $2.06 \pm 1.24$ & $4.37 \pm 0.92$\\\hline
   %R$\pm$D(COCO) & $2.67 \pm 1.5$ & $2.2 \pm 1.3$  & $4.70 \pm 0.69$\\\hline
   %T$\pm$D(COCO) & $2.52 \pm 1.42$ & $2.09 \pm 1.24$ & $4.36 \pm 0.93$\\\hline
   \end{tabular}
   }
   %}
   \end{center}
   \vspace*{-0.14in}
   \caption{\small{Sentence generation relevance (R) and thoroughness (T) human evaluation results with gold standard and \cite{karpathy2014deep} on Flickr 8k, 30k and MS-COCO datasets. D: Standard Deviation.} }
   \label{tab:sengen_res1}
   \vspace*{-0.1in}
\end{table}

\textbf{Image-Sentence Alignment Evaluation: } Similar to the experiments in \cite{karpathy2014deep,johnsoncvpr2015}, we also evaluate the image-sentence alignment quality using ranking experiments. We withhold the set of testing images and use the generated sentences as queries. 

  We process the textual query and construct $\mathcal{G}_{query}=(V_{q},E_{q})$ using the same procedure by which we construct $\mathcal{K}_b$. For each image, we take the SDG $\mathcal{G}_{img} =(V_{img},E_{img})$ and calculate similarity between the SDG and the query using the following formula:
\small{  
  \begin{equation*}
\begin{aligned}
& Sim(\mathcal{G}_{query},\mathcal{G}_{img}) = \frac{\sum_{v_q \in V_{q}} max_{v_{img} \in V_{img}}(sim(v_q,v_{img}))}{|V_q|}\\
& sim(v_q,v_{img}) = (wnsim(label(v_q),label(v_{img})) +\\ 
& Jaccard(neighbors(v_q),neighbors(v_{img})))/2.
\end{aligned}   
  \end{equation*}
  \vspace{-0.05in} 
  }

\normalsize  
Similarity between two vertices are calculated based on their word-meaning similarity and neighbor similarity. Here $wnsim(.,.)$ is WordNet-Lin Similarity \cite{lin1998information} between two words and $Jaccard(.,.)$ is the standard Jaccard coefficient similarity. Based on the above similarity measure, we give the image retrieval results compared with few of the state-of-the-art results in Table \ref{retrieval}.
\begin{table}[!htbp]
\centering
\resizebox{\columnwidth}{!}{
\small{
\begin{tabular}{|l||l|l|l|l|}
\cline{1-5}
&\multicolumn{4}{|c|}{Flickr8k}\\
\cline{1-5}
\textbf{Model} & \textbf{R@1} & \textbf{R@5} & \textbf{R@10} & \textbf{Med r}\\
\cline{1-5}
\cite{karpathy2014deep} BRNN & 11.8 &	32.1 &	44.7 & 12.4\\
\hline
Our Method-SDG & \textbf{18.1} & \textbf{39.0} & \textbf{50.0} & \textbf{10.5}\\
\cline{1-5}
&\multicolumn{4}{|c|}{Flickr30k}\\
\cline{1-5}
\cite{karpathy2014deep} BRNN & 15.2	& 37.7	& 50.5 & 9.2 \\
\hline
Our Method-SDG & \textbf{26.5} & \textbf{48.7}	& \textbf{59.4} & \textbf{6.0}\\
\hline
&\multicolumn{4}{|c|}{MS-COCO}\\
\cline{1-5}
\cite{karpathy2014deep} BRNN (1k) & \textbf{20.9}	& \textbf{52.8}	& \textbf{69.2}	& \textbf{4.0} \\
\hline
Our Method-SDG (1k) & 19.3 & 35.5	& 49.0 & 11.0\\
\hline
\hline
Our Method-SDG (2k) & 15.4 & 32.5	& 42.2 & 17.0\\
\hline
\end{tabular}
}
}
\caption{\small{Image-Search Results: We report the recall@K (for $K=1,5 \text{ and }10$) and Med r (Median Rank) metric for Flickr8k, 30k and COCO datasets. For COCO, we experimented on first 1000  (1k) and random 2000 (2k) validation images.}}
\label{retrieval}
\vspace{-0.2in}
\end{table}

  One of the primary contributions of our work is the Knowledge-Structure representation that bridges the gap between semantic information in text and images. From the results of this experiment, the benefit of having such an intermediate representation is easy to observe.
%   \begin{table}[htbp]
%   \begin{center}
%   %{\footnotesize
%   \resizebox{\columnwidth}{!}{%
%   \begin{tabular}{|p{2.0cm}||c|c|c|c|}
%   \hline 
%   Experiment & \cite{karpathy2014deep} & Our Method & Gold Standard\\ 
%   \hline 
%   Relevancy $\pm$ deviation & $1.93 \pm 1.32$ & $2.43 \pm 1.42$  & $4.78 \pm 0.61$\\\hline
%   Thoroughness $\pm$ deviation & $ 2.17 \pm 1.34$ & $2.49 \pm 1.42$ & $4.52 \pm 0.93$\\\hline
%   \end{tabular}
%   }
%   %}
%   \end{center}
%   \vspace*{-0.1in}
%   \caption{\small{Sentence generation relevancy and thoroughness human evaluation results with gold standard and \cite{karpathy2014deep} on Flickr 30k dataset.} }
%   \label{tab:sengen_res2}
%   \vspace*{-0.2in}
%   \end{table}
% Describe by examples.

\textbf{Example Images and SDGs:}  
   %According to the reported performances on large scale datasets such as the ImageNet visual recognition dataset \cite{ILSVRCarxiv14} and  the 205 scene dataset \cite{zhou2014places}, the performance of  state-of-the-art object recognition is around 40\% using average precision metric and for scene classification is 50\% reported accuracy. Thus, even the output of the state-of-the-art models on the Flickr 8K testing images is quite noisy. Even after adding the scene-constituent outputs, there were many images which produced only one object with good score. In the above experiments, we consider all testing images.
As examples, we pick a few images which produces objects and scene recognitions with comparably good confidence scores. %More specifically, we choose a few images that have at least two objects detected with high score.
  %The images are given in Figure \ref{fig:rich}. Our systems' outputs are provided for these images in Figure \ref{fig:graph_rich}.
The images and their corresponding SDGs are provided in Figure \ref{fig:graph_rich}.  As we can observe, the information produced by these SDGs are easily processed by machines. We can answer questions such as \textit{how entities interact in an event, which possible events are in the scene and how entities interact in a scene}. We should also mention that the concept-level modeling provided by SDGs is what separates this work from other recent approaches \cite{johnsoncvpr2015}. Furthermore, comparing these structures with the K-Parser output in Figure \ref{fig:kparser1}, we can see how the sentences and images can seamlessly converge to such space of graphical representations. This could have huge repercussions in search in Image and Textual space and storing knowledge from images and text together in a unified Knowledge Base.

\vspace{-0.05in}
\section{Conclusion}
\vspace{-0.05in}
This paper introduced a reasoning module to generate textual descriptions from images by first constructing a new intermediate semantic representation, namely the Scene Description Graph (SDG), which is later used to generate sentences. The reasoning module uses an automatically constructed  Knowledge Base created from text, to capture ``commonsense'' knowledge. Having built the Knowledge Base, we proposed a method of obtaining such SDGs from noisy labels using our prediction system. The SDG is a representation of the scene in view that integrates direct visual knowledge (objects and their locations in the scene) with background commonsense knowledge. In addition, the SDGs have a structure similar to  semantic representations of sentences, thus facilitating the interaction between Vision and Natural Language.  The notion of the SDG has great potential. Here we used the SDG for the automatic creation of sentences describing the scene; but, equipped with background knowledge, it also allows reasoning and question/answering about the scene \footnote{Please see appendix for an example.}.

To demonstrate the effectiveness of the sentences and constructed SDGs, we performed a number of experiments. Our AMT evaluations on popular datasets show that our sentences performs comparatively well with respect to the state-of-the-art in measures of relevance and thoroughness. A Gold-Standard based evaluation shows that our output SDGs can detect events and entities with comparable accuracy as a state-of-the-art system. And lastly, our Image Retrieval experiment shows that the Image-Sentence alignment quality is comparable with state-of-the-art results.
%For evaluations, we performed a number of experiments. In particular, we evaluated how thoroughly we can detect events and entities from an image. 
 % and we provide few examples to show the rich set of related concepts that can be produced by our system. %Our experiments and evaluations collectively demonstrates the effectiveness of the SDG structures and our approach of computing such structures from images.

%In this paper, we presented an automatic scene analysis system that integrates deep learning based visual perception module and non-domain-specific common-sense reasoning module. The output of the presented system is scene description graph. Experimental results show that the graph can be further used to do event/entity detection and generate natural language descriptions. Most importantly, several case studies show that a rich set of related concepts can be produced from our system. Along with the rich output, our algorithm intends to provide a logically sound approach to get the final SDG from the set of noisy labels.  As a result, we believe that we are able to avoid complete gibberish results. For future work, we can feed such \textit{has-Structure}s to an Answer Set Programming module as predicates and used as facts to get lot of interesting inferences. 

\newpage
{\small
\bibliographystyle{ieee}
\bibliography{vision_nlp_aaai}

\begin{thebibliography}{10}\itemsep=-1pt

\bibitem{Bengio2012}
Y.~Bengio, A.~Courville, and P.~Vincent.
\newblock Representation learning: A review and new perspectives.
\newblock {\em Pattern Analysis and Machine Intelligence, IEEE Transactions
  on}, 35(8):1798--1828, 2013.

\bibitem{Chaudhri:2007:EEB:1298406.1298435}
V.~K. Chaudhri, B.~E. John, S.~Mishra, J.~Pacheco, B.~Porter, and A.~Spaulding.
\newblock Enabling experts to build knowledge bases from science textbooks.
\newblock In {\em Proceedings of the 4th International Conference on Knowledge
  Capture}, K-CAP '07, pages 159--166, New York, NY, USA, 2007. ACM.

\bibitem{chen2014learning}
X.~Chen and C.~L. Zitnick.
\newblock Learning a recurrent visual representation for image caption
  generation.
\newblock {\em arXiv preprint arXiv:1411.5654}, 2014.

\bibitem{Clark2004}
P.~Clark, B.~Porter, and B.~P. Works.
\newblock Km-the knowledge machine 2.0: Users manual.
\newblock {\em Department of Computer Science, University of Texas at Austin},
  2004.

\bibitem{dalal2005histograms}
N.~Dalal and B.~Triggs.
\newblock Histograms of oriented gradients for human detection.
\newblock In {\em Computer Vision and Pattern Recognition, 2005. CVPR 2005.
  IEEE Computer Society Conference on}, volume~1, pages 886--893. IEEE, 2005.

\bibitem{Davis:2015:CRC:2817191.2701413}
E.~Davis and G.~Marcus.
\newblock Commonsense reasoning and commonsense knowledge in artificial
  intelligence.
\newblock {\em Commun. ACM}, 58(9):92--103, Aug. 2015.

\bibitem{DBLP:conf/cvpr/DivvalaFG14}
S.~K. Divvala, A.~Farhadi, and C.~Guestrin.
\newblock Learning everything about anything: Webly-supervised visual concept
  learning.
\newblock In {\em 2014 {IEEE} Conference on Computer Vision and Pattern
  Recognition, {CVPR} 2014, Columbus, OH, USA, June 23-28, 2014}, pages
  3270--3277, 2014.

\bibitem{donahue2014long}
J.~Donahue, L.~A. Hendricks, S.~Guadarrama, M.~Rohrbach, S.~Venugopalan,
  K.~Saenko, and T.~Darrell.
\newblock Long-term recurrent convolutional networks for visual recognition and
  description.
\newblock {\em arXiv preprint arXiv:1411.4389}, 2014.

\bibitem{donahue2014decaf}
J.~Donahue, Y.~Jia, O.~Vinyals, J.~Hoffman, N.~Zhang, E.~Tzeng, and T.~Darrell.
\newblock Decaf: A deep convolutional activation feature for generic visual
  recognition.
\newblock In {\em Proceedings of the 31st International Conference on Machine
  Learning (ICML-14)}, pages 647--655, 2014.

\bibitem{DBLP:conf/emnlp/ElliottK13}
D.~Elliott and F.~Keller.
\newblock Image description using visual dependency representations.
\newblock In {\em Proceedings of the 2013 Conference on Empirical Methods in
  Natural Language Processing, {EMNLP} 2013, 18-21 October 2013, Grand Hyatt
  Seattle, Seattle, Washington, USA, {A} meeting of SIGDAT, a Special Interest
  Group of the {ACL}}, pages 1292--1302, 2013.

\bibitem{farhadi2009describing}
A.~Farhadi, I.~Endres, D.~Hoiem, and D.~Forsyth.
\newblock Describing objects by their attributes.
\newblock In {\em Computer Vision and Pattern Recognition, 2009. CVPR 2009.
  IEEE Conference on}, pages 1778--1785. IEEE, 2009.

\bibitem{FarhadiPictureStory}
A.~Farhadi, M.~Hejrati, M.~A. Sadeghi, P.~Young, C.~Rashtchian, J.~Hockenmaier,
  and D.~Forsyth.
\newblock Every picture tells a story: Generating sentences from images.
\newblock In {\em Proceedings of the 11th European Conference on Computer
  Vision: Part IV}, ECCV'10, pages 15--29, Berlin, Heidelberg, 2010.
  Springer-Verlag.

\bibitem{felzenszwalb2008discriminatively}
P.~Felzenszwalb, D.~McAllester, and D.~Ramanan.
\newblock A discriminatively trained, multiscale, deformable part model.
\newblock In {\em Computer Vision and Pattern Recognition, 2008. CVPR 2008.
  IEEE Conference on}, pages 1--8. IEEE, 2008.

\bibitem{Gatt:2009:SRE:1610195.1610208}
A.~Gatt and E.~Reiter.
\newblock Simplenlg: A realisation engine for practical applications.
\newblock In {\em Proceedings of the 12th European Workshop on Natural Language
  Generation}, ENLG '09, pages 90--93, Stroudsburg, PA, USA, 2009. Association
  for Computational Linguistics.

\bibitem{girshick14CVPR}
R.~Girshick, J.~Donahue, T.~Darrell, and J.~Malik.
\newblock Rich feature hierarchies for accurate object detection and semantic
  segmentation.
\newblock In {\em Computer Vision and Pattern Recognition}, 2014.

\bibitem{Gunning_projecthalo}
D.~Gunning, V.~K. Chaudhri, P.~Clark, K.~Barker, S.~yi~Chaw, B.~Grosof,
  A.~Leung, D.~Mcdonald, S.~Mishra, J.~Pacheco, B.~Porter, A.~Spaulding,
  D.~Tecuci, and J.~Tien.
\newblock Project halo update—progress toward digital aristotle.

\bibitem{gupta2007objects}
A.~Gupta and L.~S. Davis.
\newblock Objects in action: An approach for combining action understanding and
  object perception.
\newblock In {\em Computer Vision and Pattern Recognition, 2007. CVPR'07. IEEE
  Conference on}, pages 1--8. IEEE, 2007.

\bibitem{havasi2007conceptnet}
C.~Havasi, R.~Speer, and J.~Alonso.
\newblock Conceptnet 3: a flexible, multilingual semantic network for common
  sense knowledge.
\newblock In {\em Recent advances in natural language processing}, pages
  27--29. Citeseer, 2007.

\bibitem{hodosh2013framing}
M.~Hodosh, P.~Young, and J.~Hockenmaier.
\newblock Framing image description as a ranking task: Data, models and
  evaluation metrics.
\newblock {\em Journal of Artificial Intelligence Research}, pages 853--899,
  2013.

\bibitem{johnsoncvpr2015}
J.~Johnson, R.~Krishna, M.~Stark, J.~Li, M.~Bernstein, and L.~Fei-Fei.
\newblock Image retrieval using scene graphs.
\newblock In {\em IEEE Conference on Computer Vision and Pattern Recognition
  (CVPR)}, June 2015.

\bibitem{karpathy2014deep}
A.~Karpathy and F.-F. Li.
\newblock Deep visual-semantic alignments for generating image descriptions.
\newblock {\em arXiv preprint arXiv:1412.2306}, 2014.

\bibitem{kiros2014unifying}
R.~Kiros, R.~Salakhutdinov, and R.~S. Zemel.
\newblock Unifying visual-semantic embeddings with multimodal neural language
  models.
\newblock {\em arXiv preprint arXiv:1411.2539}, 2014.

\bibitem{Krizhevsky2012}
A.~Krizhevsky, I.~Sutskever, and G.~Hinton.
\newblock Imagenet classification with deep convolutional neural networks.
\newblock In {\em NIPS 2012}, 2013.

\bibitem{Kulkarni11babytalk:}
G.~Kulkarni, V.~Premraj, S.~Dhar, S.~Li, Y.~Choi, A.~C. Berg, and T.~L. Berg.
\newblock Baby talk: Understanding and generating image descriptions.
\newblock In {\em Proceedings of the 24th CVPR}, 2011.

\bibitem{Kuznetsova:2012:CGN:2390524.2390575}
P.~Kuznetsova, V.~Ordonez, A.~C. Berg, T.~L. Berg, and Y.~Choi.
\newblock Collective generation of natural image descriptions.
\newblock In {\em Proceedings of the 50th Annual Meeting of the Association for
  Computational Linguistics: Long Papers - Volume 1}, ACL '12, pages 359--368,
  Stroudsburg, PA, USA, 2012. Association for Computational Linguistics.

\bibitem{lampert2009learning}
C.~H. Lampert, H.~Nickisch, and S.~Harmeling.
\newblock Learning to detect unseen object classes by between-class attribute
  transfer.
\newblock In {\em Computer Vision and Pattern Recognition, 2009. CVPR 2009.
  IEEE Conference on}, pages 951--958. IEEE, 2009.

\bibitem{laptev2005space}
I.~Laptev.
\newblock On space-time interest points.
\newblock {\em International Journal of Computer Vision}, 64(2-3):107--123,
  2005.

\bibitem{Lenat:1995:CLI:219717.219745}
D.~B. Lenat.
\newblock Cyc: A large-scale investment in knowledge infrastructure.
\newblock {\em Commun. ACM}, 38(11):33--38, Nov. 1995.

\bibitem{lin1998information}
D.~Lin.
\newblock An information-theoretic definition of similarity.
\newblock In {\em ICML}, volume~98, pages 296--304, 1998.

\bibitem{lowe1999object}
D.~G. Lowe.
\newblock Object recognition from local scale-invariant features.
\newblock In {\em Computer vision, 1999. The proceedings of the seventh IEEE
  international conference on}, volume~2, pages 1150--1157. Ieee, 1999.

\bibitem{mao2014explain}
J.~Mao, W.~Xu, Y.~Yang, J.~Wang, and A.~L. Yuille.
\newblock Explain images with multimodal recurrent neural networks.
\newblock {\em arXiv preprint arXiv:1410.1090}, 2014.

\bibitem{messing2009activity}
R.~Messing, C.~Pal, and H.~Kautz.
\newblock Activity recognition using the velocity histories of tracked
  keypoints.
\newblock In {\em Computer Vision, 2009 IEEE 12th International Conference on},
  pages 104--111. IEEE, 2009.

\bibitem{conf/eccv/OgaleKA06}
A.~S. Ogale, A.~Karapurkar, and Y.~Aloimonos.
\newblock View-invariant modeling and recognition of human actions using
  grammars.
\newblock In R.~Vidal, A.~Heyden, and Y.~Ma, editors, {\em WDV}, volume 4358 of
  {\em Lecture Notes in Computer Science}, pages 115--126. Springer, 2006.

\bibitem{im2textOrdonez}
V.~Ordonez, G.~Kulkarni, and T.~L. Berg.
\newblock Im2text: Describing images using 1 million captioned photographs.
\newblock In J.~Shawe-Taylor, R.~S. Zemel, P.~L. Bartlett, F.~C.~N. Pereira,
  and K.~Q. Weinberger, editors, {\em NIPS}, pages 1143--1151, 2011.

\bibitem{razavian2014cnn}
A.~S. Razavian, H.~Azizpour, J.~Sullivan, and S.~Carlsson.
\newblock Cnn features off-the-shelf: an astounding baseline for recognition.
\newblock In {\em Computer Vision and Pattern Recognition Workshops (CVPRW),
  2014 IEEE Conference on}, pages 512--519. IEEE, 2014.

\bibitem{Santofimia12CommonSenseVision}
M.~Santofimia, J.~Martinez-del Rincon, and J.-C. Nebel.
\newblock {Common-Sense Knowledge for a Computer Vision System for Human Action
  Recognition}.
\newblock In J.~Bravo, R.~Herv\'{a}s, and M.~Rodr\'{\i}guez, editors, {\em
  Ambient Assisted Living and Home Care}, volume 7657 of {\em Lecture Notes in
  Computer Science}, pages 159--166. Springer Berlin Heidelberg, 2012.

\bibitem{schuster-EtAl:2015:VL}
S.~Schuster, R.~Krishna, A.~Chang, L.~Fei-Fei, and C.~D. Manning.
\newblock Generating semantically precise scene graphs from textual
  descriptions for improved image retrieval.
\newblock In {\em Proceedings of the Fourth Workshop on Vision and Language},
  pages 70--80, Lisbon, Portugal, September 2015. Association for Computational
  Linguistics.

\bibitem{bnlearn}
M.~Scutari.
\newblock Learning bayesian networks with the {bnlearn} {R} package.
\newblock {\em Journal of Statistical Software}, 35(3):1--22, 2010.

\bibitem{DBLP:conf/ijcai/SharmaVAB15}
A.~Sharma, N.~H. Vo, S.~Aditya, and C.~Baral.
\newblock Towards addressing the winograd schema challenge - building and using
  a semantic parser and a knowledge hunting module.
\newblock In {\em Proceedings of the Twenty-Fourth International Joint
  Conference on Artificial Intelligence, {IJCAI} 2015, Buenos Aires, Argentina,
  July 25-31, 2015}, pages 1319--1325, 2015.

\bibitem{DBLP:journals/tacl/SocherKLMN14}
R.~Socher, A.~Karpathy, Q.~V. Le, C.~D. Manning, and A.~Y. Ng.
\newblock Grounded compositional semantics for finding and describing images
  with sentences.
\newblock {\em {TACL}}, 2:207--218, 2014.

\bibitem{stork1998hal}
D.~G. Stork.
\newblock {\em HAL's Legacy: 2001's Computer as Dream and Reality}.
\newblock MIT Press, 1998.

\bibitem{teo2015gestaltist}
C.~L. Teo, C.~Ferm{\"u}ller, and Y.~Aloimonos.
\newblock A gestaltist approach to contour-based object recognition: Combining
  bottom-up and top-down cues.
\newblock {\em The International Journal of Robotics Research}, page
  0278364914558493, 2015.

\bibitem{teo2013embedding}
C.~L. Teo, A.~Myers, C.~Fermuller, and Y.~Aloimonos.
\newblock Embedding high-level information into low level vision: Efficient
  object search in clutter.
\newblock In {\em Robotics and Automation (ICRA), 2013 IEEE International
  Conference on}, pages 126--132. IEEE, 2013.

\bibitem{vinyals2014show}
O.~Vinyals, A.~Toshev, S.~Bengio, and D.~Erhan.
\newblock Show and tell: A neural image caption generator.
\newblock {\em arXiv preprint arXiv:1411.4555}, 2014.

\bibitem{wang2011action}
H.~Wang, A.~Klaser, C.~Schmid, and C.-L. Liu.
\newblock Action recognition by dense trajectories.
\newblock In {\em Computer Vision and Pattern Recognition (CVPR), 2011 IEEE
  Conference on}, pages 3169--3176. IEEE, 2011.

\bibitem{yang2014cognitive}
Y.~Yang, C.~Ferm{\"u}ller, Y.~Aloimonos, and A.~Guha.
\newblock A cognitive system for understanding human manipulation actions.
\newblock {\em Advances in Cognitive Systems}, 3:67--86, 2014.

\bibitem{Yang:2011:CSG:2145432.2145484}
Y.~Yang, C.~L. Teo, H.~Daum{\'e}, III, and Y.~Aloimonos.
\newblock Corpus-guided sentence generation of natural images.
\newblock In {\em Proceedings of the Conference on Empirical Methods in Natural
  Language Processing}, EMNLP '11, pages 444--454, Stroudsburg, PA, USA, 2011.
  Association for Computational Linguistics.

\bibitem{journals/pieee/YaoYLLZ10}
B.~Z. Yao, X.~Yang, L.~Lin, M.~W. Lee, and S.~C. Zhu.
\newblock I2t: Image parsing to text description.
\newblock {\em Proceedings of the IEEE}, 98(8):1485--1508, 2010.

\bibitem{yu2010attribute}
X.~Yu and Y.~Aloimonos.
\newblock Attribute-based transfer learning for object categorization with
  zero/one training example.
\newblock In {\em Computer Vision--ECCV 2010}, pages 127--140. Springer Berlin
  Heidelberg, 2010.

\bibitem{yu2011active}
X.~Yu, C.~Fermuller, C.~L. Teo, Y.~Yang, and Y.~Aloimonos.
\newblock Active scene recognition with vision and language.
\newblock In {\em Computer Vision (ICCV), 2011 IEEE International Conference
  on}, pages 810--817. IEEE, 2011.

\bibitem{zhou2014places}
B.~Zhou, A.~Lapedriza, J.~Xiao, A.~Torralba, and A.~Oliva.
\newblock Learning deep features for scene recognition using places database.
\newblock {\em NIPS}, 2014.

\bibitem{conf/cvpr/ZitnickP13}
C.~L. Zitnick and D.~Parikh.
\newblock Bringing semantics into focus using visual abstraction.
\newblock In {\em CVPR}, pages 3009--3016. IEEE, 2013.

\end{thebibliography}
}



\section*{Supplementary material}

# APPENDIX of "From Images to Sentences through Scene Description Graphs using Commonsense Reasoning and Knowledge"

## Contents

# 1 Question-Answering based on SDGs

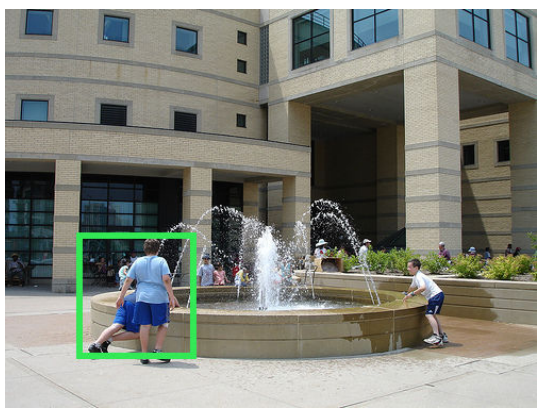

Figure 1: An Example Image from Flickr 8k: The green box shows a person is detected with score 1.36056. In this image, this is the only one instance of class "person" detected with high confidence.

For the image in Figure 1, the generated facts in SDG are:

```
has(scene,component,water).
has(scene,component,water_droplets).
has(scene,component,exterior_of_building).
has(person1,semantic_role,drinker).
has(water,semantic_role,liquid).
has(person1,semantic_role,creator).
has(drink,recipient,water).
has(drink,agent,person1).
has(drink,origin,fountain).
has(drink,next_event,make).
```

One of the advantage of SDG is, we can directly use the above triples as facts and as an input to an Answer Set Program. Now, if we pose the question that "Is someone drinking from the fountain?" in ASP, we can answer based on the above facts using the following program in Clingo-3:

```
entity(person;dog;water;shorts;frisbee).
animate(person;dog).

inanimate(A) :- not animate(A), entity(A).

drink_yes :- animate(A), has(drink,agent,A), has(drink,recipient,water).
yes_fountain(A) :- drink_yes, has(drink,agent,A), has(drink,origin, fountain).

#hide.
#show yes_fountain/1.
```

We execute the above program in Clingo-3, we get the answer as *yes_fountain(person1)*.

actual output:

## 2  Images, Corresponding SDGs and Generated Sentences

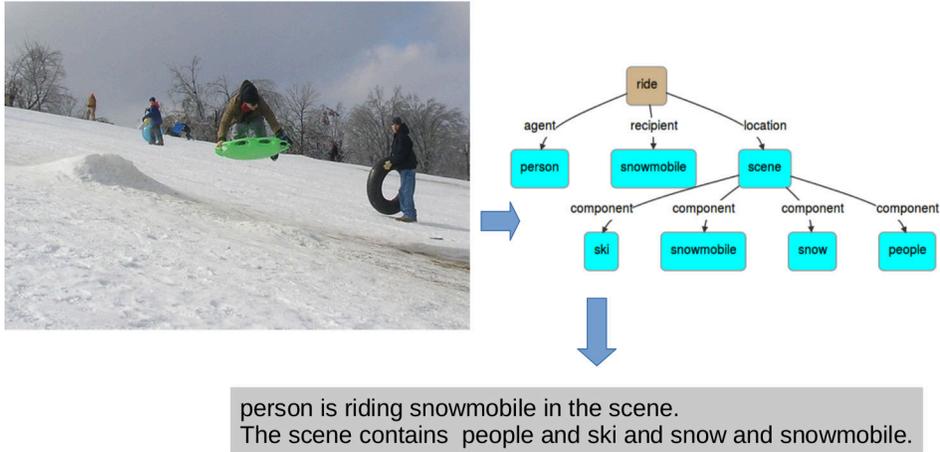

person is riding snowmobile in the scene.
The scene contains people and ski and snow and snowmobile.

Figure 2: Case Study-I

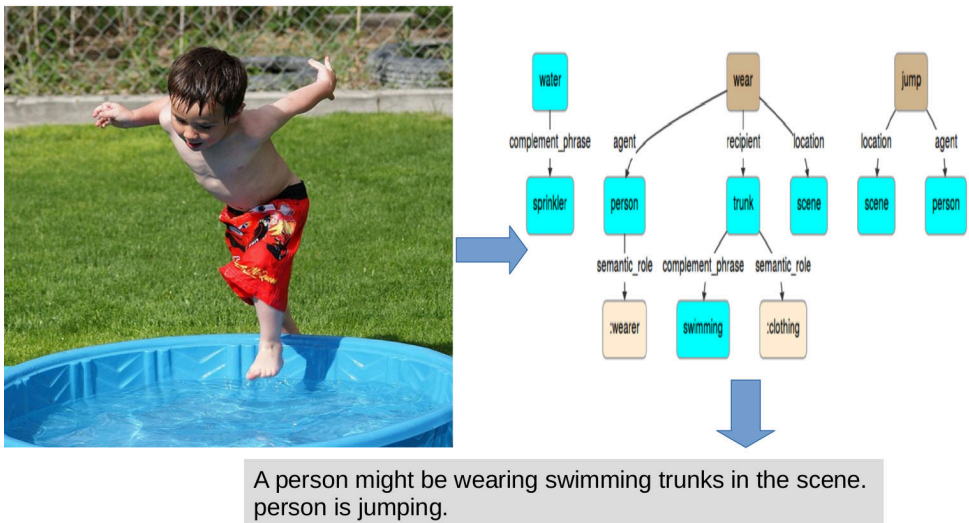

A person might be wearing swimming trunks in the scene.
person is jumping.

Figure 3: Case Study-II

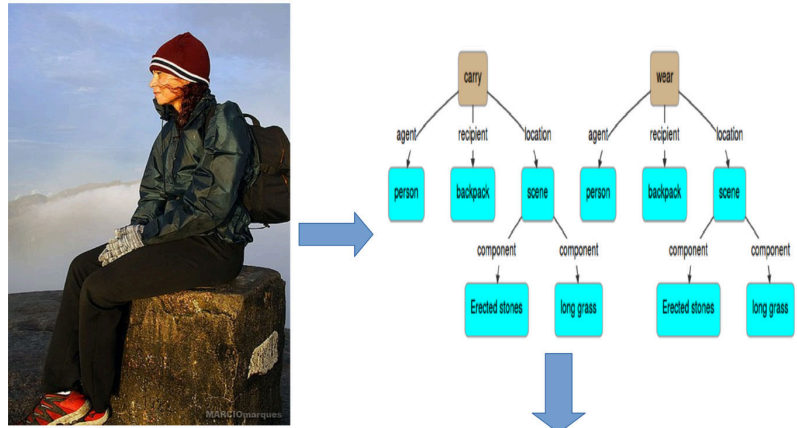

Figure 4: Case Study-III

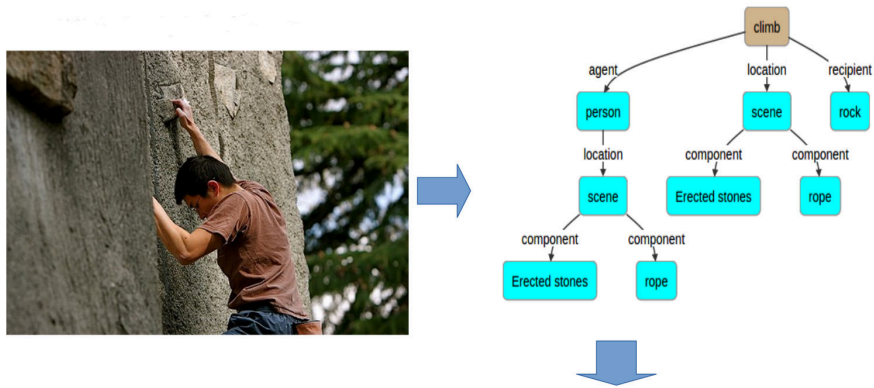

Figure 5: Case Study-IV

4

Figure 6: Case Study-V

# 3   Some More Images and Generated Sentences

person is driving car in the scene.
The scene contains street and people walk and concrete roads and booths and vehicles.

person is surfing at ocean.The scene contains large waterbody and sand and water and ocean.

Figure 7: Sentence Examples

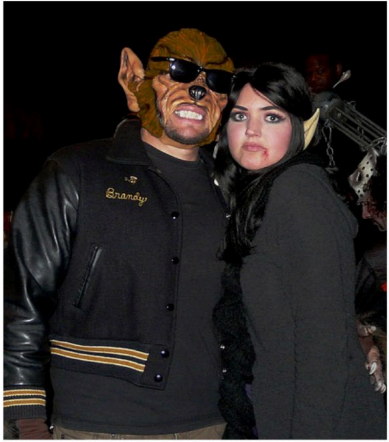
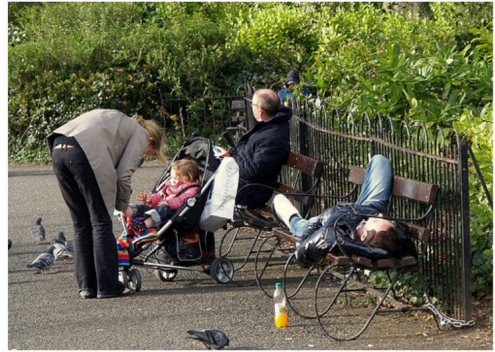

A person might be wearing sunglass in the scene.
The scene contains big size recording instruments and people wearing headphones.

person is pulling cart in the scene.
person is pushing person at cart.

Figure 8: Sentence Examples

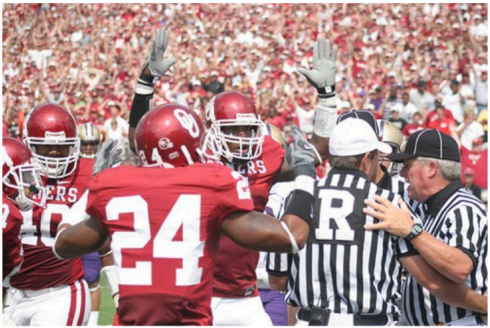
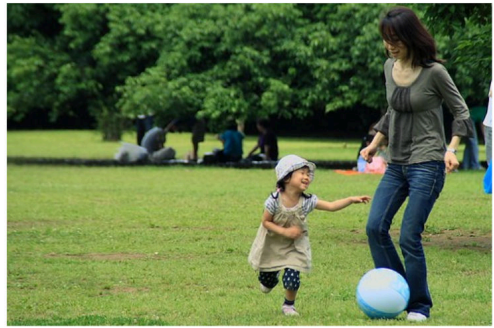

The scene contains hardwood floor and people in sportswear and crowd.

person is kicking ball in the scene.
person is hitting ball in the scene.

Figure 9: Sentence Examples

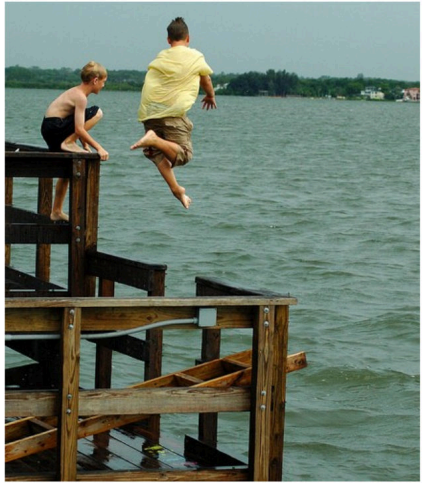 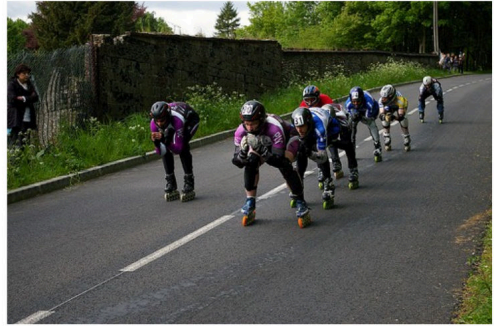

person is sitting table in the scene. person is jumping in the scene.
The scene contains water and people walk on road and large waterbody and people walk and sand.

The scene contains trees and grasses and concrete roads and artificial path and vehicles.

Figure 10: Sentence Examples

# 4  AMT Instructions

**Instructions**

The description is rated on one scale: Correctness
- Descriptions that correctly describes the image content with higher precision have better correctness ratings.
- Score 1: The description has no relevance to the image;
- Score 2: The description have only weak relevance to the image;
- Score 3: The description have some relevance to the image;
- Score 4: The description relates closely to the image;
- Score 5: The description relates perfectly to the image.

**Please judge all four descriptions, otherwise the response is NOT valid.**

Figure 11: Relevance Evaluation Task Instructions as provided to the Turkers

**Instructions**

The description is rated on one scale: Thoroughness
- Descriptions that cover more aspects of the image content have better thoroughness ratings.
- NOTE: you are NOT asked to judge the correctness of the description.
- Score 1: The description covers noting to the image;
- Score 2: The description covers minor aspect to the image;
- Score 3: The description covers some aspects to the image;
- Score 4: The description covers many aspects to the image;
- Score 5: The description covers almost every aspect to the image.

**Please judge all four descriptions, otherwise the response is NOT valid.**

Figure 12: Thoroughness Evaluation Task Instructions as provided to the Turkers



\end{document}